\documentclass[conference]{IEEEtran}
\IEEEoverridecommandlockouts
\usepackage{cite}
\usepackage{amsmath,amssymb,amsfonts}
\usepackage{algorithmic}
\usepackage{graphicx}
\usepackage{textcomp}
\usepackage{xcolor}
\def\BibTeX{{\rm B\kern-.05em{\sc i\kern-.025em b}\kern-.08em
    T\kern-.1667em\lower.7ex\hbox{E}\kern-.125emX}}
\usepackage{color}
\usepackage{hyperref}
\hypersetup{
 colorlinks=True}
\usepackage{subfigure}
\usepackage{amsmath,amssymb,amsfonts}

\usepackage{algorithm}
\usepackage{algorithmic}

\usepackage{booktabs}
\usepackage{multirow}
\usepackage{wrapfig}
    
\begin{document}

\title{Heterogeneous Domain Adaptation with Positive and Unlabeled Data
}

\author{\IEEEauthorblockN{Junki Mori}
\IEEEauthorblockA{
\textit{NEC Corporation}\\
Kanagawa, Japan \\
junki.mori@nec.com}
\and
\IEEEauthorblockN{Ryo Furukawa}
\IEEEauthorblockA{
\textit{NEC Corporation}\\
Kanagawa, Japan \\
rfurukawa@nec.com}
\and
\IEEEauthorblockN{Isamu Teranishi}
\IEEEauthorblockA{
\textit{NEC Corporation}\\
Kanagawa, Japan \\
teranisi@nec.com}
\and
\IEEEauthorblockN{Jun Sakuma}
\IEEEauthorblockA{
\textit{Tokyo Institute of Technology}\\
\textit{RIKEN Center for Advanced Intelligence Project}\\
Tokyo, Japan \\
sakuma@c.titech.ac.jp}
}

\maketitle

\begin{abstract}
Heterogeneous unsupervised domain adaptation (HUDA) is the most challenging domain adaptation setting where the feature spaces of source and target domains are heterogeneous, and the target domain has only unlabeled data. Existing HUDA methods assume that both positive and negative examples are available in the source domain, which may not be satisfied in some real applications. This paper addresses a new challenging setting called \textit{positive and unlabeled heterogeneous unsupervised domain adaptation} (PU-HUDA), a HUDA setting where the source domain only has positives. PU-HUDA can also be viewed as an extension of PU learning where the positive and unlabeled examples are sampled from different domains. A naive combination of existing HUDA and PU learning methods is ineffective in PU-HUDA due to the gap in label distribution between the source and target domains. To overcome this issue, we propose a novel method, \textit{predictive adversarial domain adaptation} (PADA), which can predict likely positive examples from the unlabeled target data and simultaneously align the feature spaces to reduce the distribution divergence between the whole source data and the likely positive target data. PADA achieves this by a unified adversarial training framework for learning a classifier to predict positive examples and a feature transformer to transform the target feature space to that of the source. Specifically, they are both trained to fool a common discriminator that determines whether the likely positive examples are from the target or source domain. We experimentally show that PADA outperforms several baseline methods, such as the naive combination of HUDA and PU learning.

\end{abstract}

\begin{IEEEkeywords}
heterogeneous domain adaptation, PU learning, adversarial training
\end{IEEEkeywords}

\section{Introduction}
\label{introduction}

\begin{figure}[t]
    \centering
    \subfigure[Conventional HUDA setting.]{\includegraphics[width=0.4\textwidth]{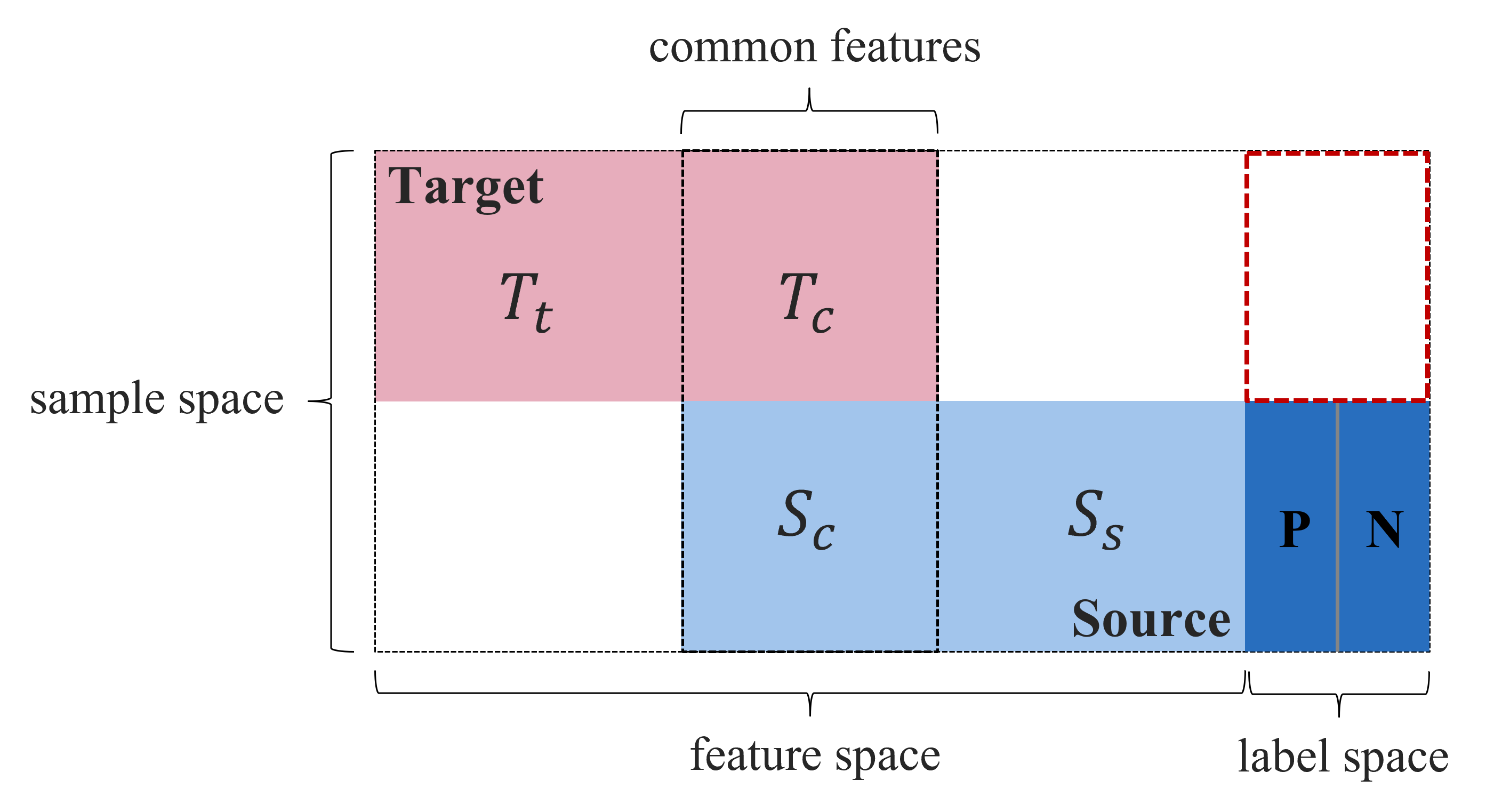}\label{fig:hda}}
    \subfigure[New PU-HUDA setting.]{\includegraphics[width=0.4\textwidth]{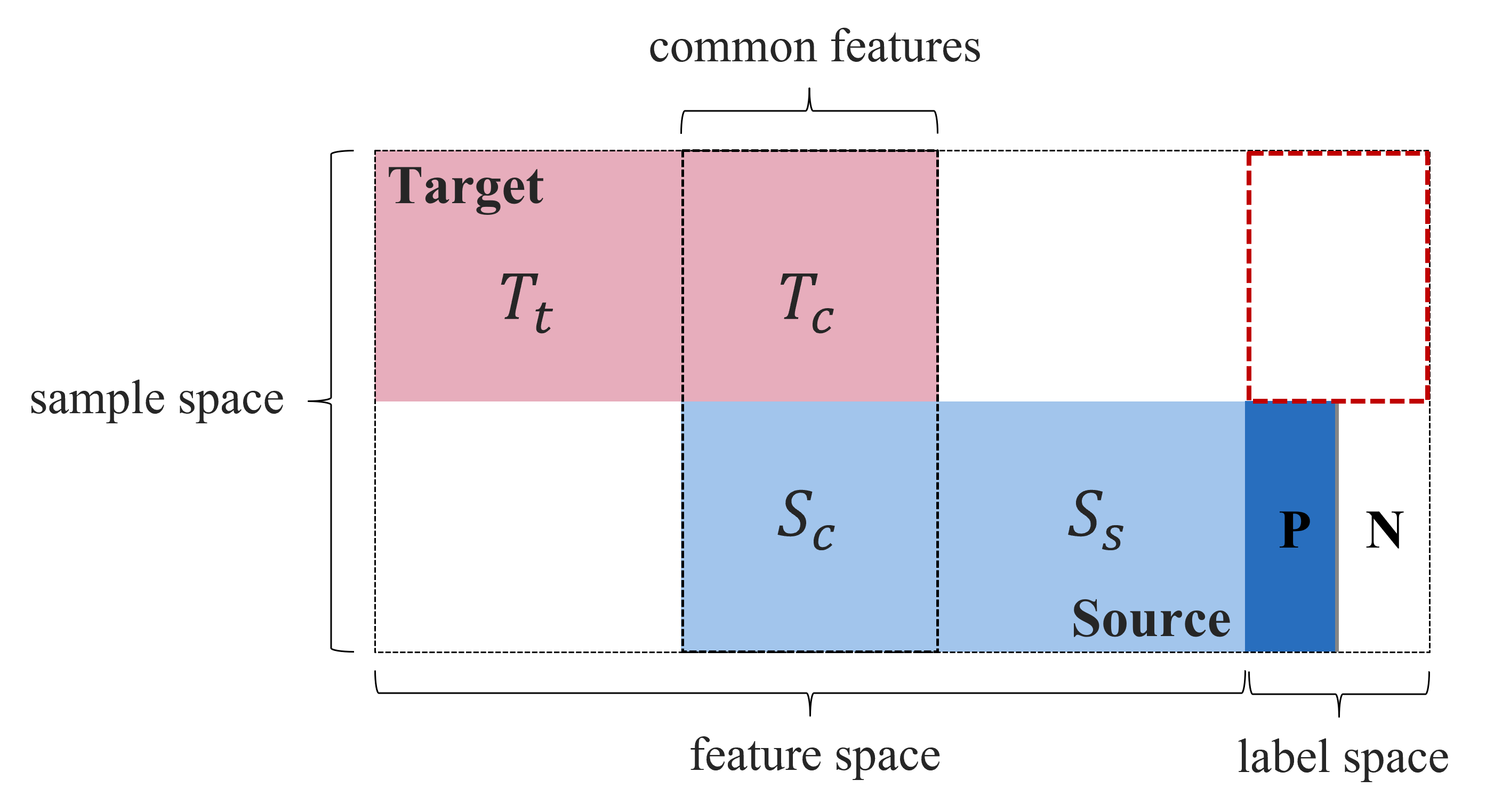}\label{fig:pu-hda}}
    \caption{View of the data structure in two HUDA settings, (a) conventional one and (b) new one where the source domain has only positive data. Here, both of them assume that the two domains share some features.} \label{fig:setting}
\end{figure}

Applying machine learning models trained with a domain to different domains may lead to performance degradation \cite{bias}. This motivates researchers to study \textit{unsupervised domain adaptation} (UDA) \cite{transfer,deepda}, which transfers knowledge from a known domain (source) where a sufficiently large number of labeled samples are available to an unknown domain (target) where only unlabeled samples are available. The UDA setting supposes that the data distributions in the source and target domains are different, but the feature space is common. However, when the data is sampled from different domains, the feature space itself often differs in real-world problems (e.g., different attributes are observed), as we show examples later. \textit{Heterogeneous unsupervised domain adaptation} (HUDA) \cite{hda} is a new UDA task that addresses different feature spaces. Existing HUDA studies \cite{kcca,hhtl,dsft,ot,sfer,glg} only consider situations where both positive and negative examples in a binary classification task are available in the source domain, as illustrated in Fig.~\ref{fig:setting} (a). However, in some real-world applications, such as the example below, the source domain may have access to only positive examples.

\textit{Estimation of potential customers in different industries.}
Consider a business alliance of a bank and a jewelry store. The goal of them is to find potential customers for the jewelry store among the bank's large customer base. This task can be considered a simple classification problem of predicting whether a given customer in the bank's database will purchase jewelry or not. To address this task, they use their own customer data represented by the following features. First, they both might possess basic demographics such as age and gender. On the other hand, each industry has their own features. The bank has detailed economic conditions represented by deposit balances, and the jewelry store has past purchase histories. Hence, this task can be considered a HUDA task with the jewelry store as the source and the bank as the target (see Fig.~\ref{fig:setting}). However, while the bank (target) do not have information on purchases of jewelry, i.e., only has unlabeled customers, unlike a conventional HUDA setting, the jewelry store (source) has only customers who have already purchased or visited the jewelry store, i.e., positive labeled customers.  

\begin{figure}[t]
    \centering
    \includegraphics[width=0.5\textwidth]{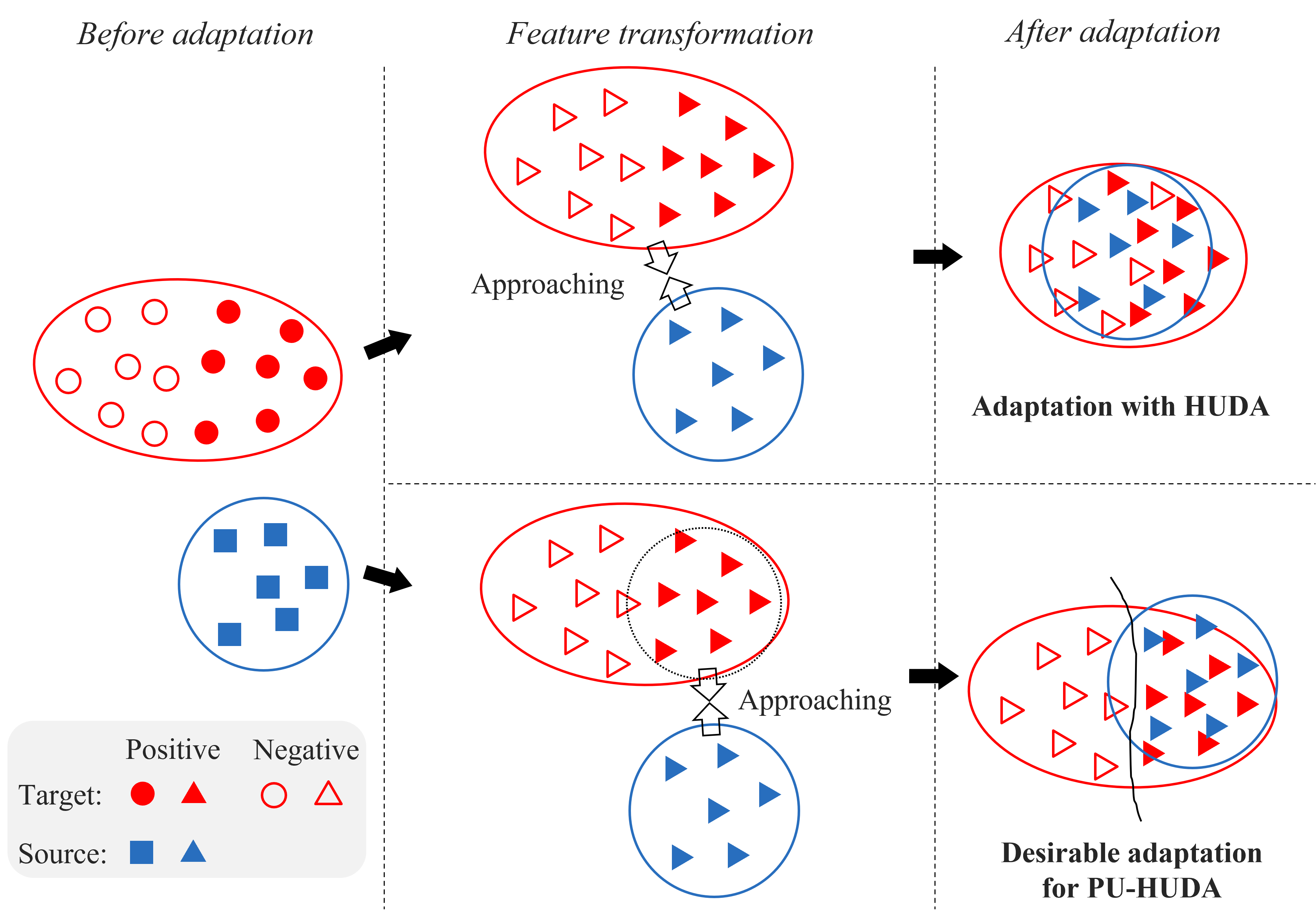}
    \caption{Comparison of adaptation mechanisms in existing HUDA methods (top) and our proposed PU-HUDA method (bottom).} \label{fig:transformation}
\end{figure}

Motivated by the above application, this paper addresses a new challenging HUDA task called \textit{positive and unlabeled heterogeneous unsupervised domain adaptation} (PU-HUDA), which consists of a source domain with positive data a and target domain with unlabeled data in the different feature spaces as described in Fig.~\ref{fig:setting} (b). PU-HUDA can be viewed as a hybrid setup of positive unlabeled (PU) learning \cite{pu}, which trains binary classifiers from positive and unlabeled data only, and HUDA. It appears that simply combining existing HUDA and PU learning methods can properly solve PU-HUDA, but it does not work well, as we demonstrate later in our experiments. This is because applying existing HUDA methods directly to PU-HUDA minimizes the divergence between the distributions of the whole source data with positive labels only and the whole target data with positive and negative labels on the transformed feature space, resulting in a less label-discriminative feature space on the target domain as shown in Fig.~\ref{fig:transformation} (top). To solve PU-HUDA properly, we need to minimize the distribution divergence between the whole source data and the target data with only positive labels on the transformed feature space as described in Fig.~\ref{fig:transformation} (bottom).


To address the above issue, we regard PU-HUDA as a task to identify likely positive examples from target data and reduce the distribution divergence between the whole source examples and the target examples likely to be positive. To solve this task, we propose a new method using adversarial training as in Generative Neural Network (GAN) \cite{gan}. Our proposed method mainly consists of three models, (1) a feature transformer to transform the target feature space to that of the source, (2) a classifier to predict likely positive examples from the unlabeled target data transformed by the feature transformer, and (3) a discriminator to determine whether the likely positive examples transformed by the feature transformer are from the target or source domain. The classifier is trained to identify the likely positive examples indistinguishable from the source examples by the discriminator. The feature transformer is trained to transform the target examples predicted as positive by the classifier into the indistinguishable examples from the source examples by the discriminator. To achieve these training, we adopt an objective function based
on Kullback-Leibler (KL) divergence inspired by Predictive Adversarial Networks (PAN) \cite{pan}, a recently proposed PU learning method. We call this approach \textit{predictive adversarial domain adaptation} (PADA). PADA can make the positive target and the whole source data closer while keeping the negative target and the whole source data away, as shown in Fig.~\ref{fig:transformation} (bottom). As a result, PADA obtains a classifier that provides good classification accuracy on the feature space transformed by a feature transformer.

Our contributions can be summarized as follows:
\begin{itemize}
    \item We define a new domain adaptation task, \textit{positive and unlabeled heterogeneous unsupervised domain adaptation} (PU-HUDA). To the best of our knowledge, this paper is the first study to address PU-HUDA.
    \item We propose a novel method using adversarial training, \textit{positive-adversarial domain adaptation} (PADA). The PADA's objective is based on KL divergence inspired by PAN \cite{pan}.
    \item We experimentally show that PADA outperforms several baseline methods to solve PU-HUDA.
\end{itemize}

\section{Related Works}
We review previous studies related to the topic of this paper and discuss how different our work is from them.

\begin{table*}[t]
\centering
\caption{Setting comparison of this paper with existing HUDA methods.}\label{huda_comparison}
\begin{tabular}{cccc}
\hline
\, \multirow{2}{*}{Method} \, & \, Some common samples \, & \, Some common features \, & \,  Source negative examples \\
 & are necessary? & are necessary? & are necessary? \\
\hline
KCCA \cite{kcca} & Yes & \textbf{No} & Yes \\
HHTL \cite{hhtl} & Yes & \textbf{No} & Yes \\
DSFT \cite{dsft} & \textbf{No} & Yes & Yes \\
OT-based \cite{ot} & \textbf{No} & Yes & Yes \\
SFER \cite{sfer} & \textbf{No} & \textbf{No} & Yes \\
GLG \cite{glg} & \textbf{No} & \textbf{No} & Yes \\
\textbf{PADA(this paper)} & \textbf{No} & Yes & \textbf{No} \\
\hline
\end{tabular}
\end{table*}

\subsection{Heterogeneous Unsupervised Domain Adaptation}
The common idea of existing HUDA techniques is to learn feature transformers to transform the source and target data into a homogeneous feature space to minimize the distribution distance of the two domains on that feature space. Due to this difficulty of dealing with two heterogeneous feature spaces, most HUDA techniques require some supplementary information to bridge the two domains. For example, KCCA \cite{kcca} and HHTL \cite{hhtl} require paired (common) samples between the two domains, which are often difficult to obtain in real cases. DSFT \cite{dsft} and OT(optimal transport)-based method \cite{ot} require common features between the two domains as in this paper\footnote{While DSFT and this paper consider the situation where the source and target each have domain-specific features, OT-based method considers the situation where only the target has domain-specific features.}. CL-SCL \cite{pivot} and FSR \cite{fsr} require semantically equivalent word pairs and meta-features, respectively. There are some studies that do not explicitly require supplementary information, such as GLG \cite{glg} and SFER \cite{sfer} although they implicitly assume that the two domains have sufficiently similar features. All of the methods listed here require both positive and negative examples in the source domain. A comparison of this paper with these methods is summarized in Table~\ref{huda_comparison}.

\subsection{PU learning}
PU learning trains a binary classification from positive and unlabeled examples without labeled negative examples \cite{pu}. Early PU learning methods adopt the two-step technique that first identifies reliable negative examples and then conducts supervised learning \cite{two-steps1,two-steps2}. Some studies propose PU learning methods by considering unlabeled data as negative data having label noise \cite{noise1,noise2}. Another promising branch of PU learning employs the framework of cost-sensitive learning, such as uPU \cite{upu} and nnPU \cite{nnpu}. Recently, Hu et al. \cite{pan} proposed the state-of-the-art PU learning method, Predictive Adversarial Networks (PAN) based on the revised architecture of GAN. Besides performance, PAN has the advantage of not requiring class prior probability.

\subsection{Domain Adaptation in PU learning setting}
As in our study, there are several recent studies that address domain adaptation in the context of PU learning. For example, Sonntag et al. \cite{puda} considers the domain adaptation scenario where the source domain has labeled data of all classes, and the target domain has unlabeled data and a few positive labeled data in the homogeneous feature space. 
The setting differs from ours which is more difficult, in that the target domain has positive labeled data and the source domain has both positive and negative data, and in the homogeneity of the feature space.
The scenario where both target and source domains have unlabeled and positive labeled data is also treated for a link prediction task in the homogeneous setting \cite{plt1} and heterogeneous setting \cite{plt2}, respectively. 
Their settings are also different from ours where the source domain has only positive labeled data and the target domain has only unlabeled data.
In addition, there are studies that address open set domain adaptation (OSDA) by considering it as PU learning \cite{osda1,osda2}. OSDA performs domain adaptation while also rejects target classes that are not present in the source domain as unknown. In their settings, the source domain has a positive (known) labeled data and the target domain has unlabeled data including positive and negative (unknown) data. However, they consider the homogeneous feature space unlike our study. Thus far, no study has examined the PU learning task in the HUDA setting.

\begin{figure*}[th]
    \centering
    \includegraphics[width=0.9\textwidth]{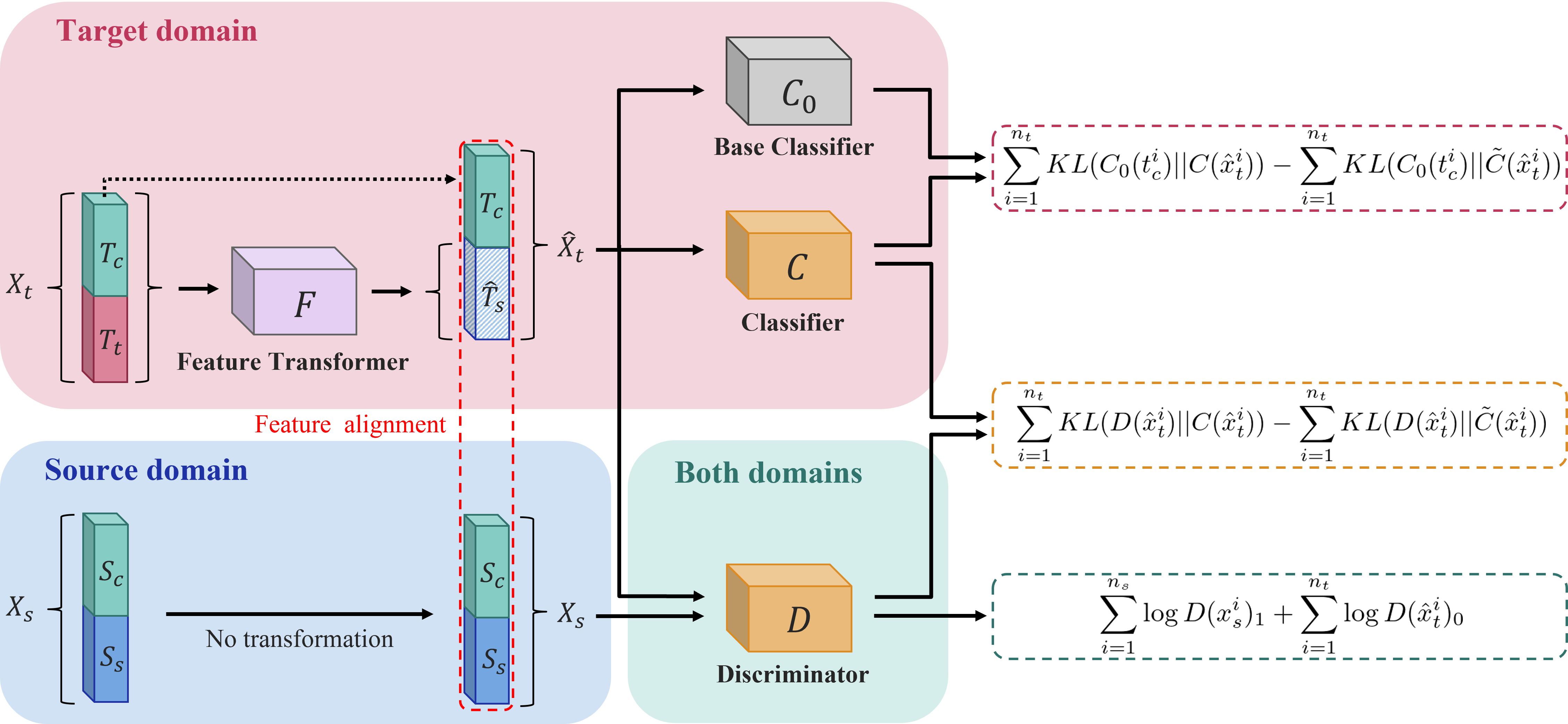}
    \caption{Overview of the proposed method, PADA using soft-labeling mechanism.} \label{fig:proposal}
\end{figure*}

\section{Positive and Unlabeled Heterogeneous Domain Adaptation}
\subsection{Problem Definition}
\label{problem}

We define a problem addressed in this paper, PU-HUDA. We suppose that the source domain and target domain share some common features as in DSFT \cite{dsft} (although our experimental results show that our proposed method works even when the number of common features is sufficiently small).  Let $X_s=[S_c;S_s] \in \mathbb{R}^{n_s\times c}\times\mathbb{R}^{n_s\times s}$  and $X_t=[T_c;T_t] \in \mathbb{R}^{n_t\times c} \times \mathbb{R}^{n_t\times t}$ be the source data matrix with $n_s$ data samples and target data matrix with $n_t$ data samples, respectively. Here, $S_c$ and $T_c$ are the data matrices in the common feature space with $c$ dimension in the source and target domain, respectively. $S_s$ and $T_t$ are the data matrices in the source and target specific feature spaces with $s$ and $t$ dimensions, respectively. We suppose that all data in the source domain is labeled as positive while all data in the target domain is unlabeled (including both positive and negative examples). We remark that this setting is different from the conventional HUDA setting where the source domain has both positive and negative labeled data. The main tasks of the conventional HUDA and PU-HUDA are the same, i.e., to predict the binary labels of the target data. These settings are visualized in Fig.~\ref{fig:setting}. 

The simplest baseline for solving PU-HUDA is to use only common features $S_c$ and $T_c$. Existing PU learning methods can be directly applied to the common features. However, since existing PU learning methods cannot use domain-specific features, this baseline method provides only limited performance. In solving PU-HUDA, PU learning that can take advantage of domain specific features in the heterogeneous setting is essential.

\subsection{Difficulty of PU-HUDA}
\label{difficulty}
One naive solution to tackle PU-HUDA is to directly combine the existing HUDA and PU learning methods. In this naive combination approach, we first train feature transformers that transform $X_s$ and $X_t$ into $\hat{X}_s$ and $\hat{X}_t$ in a new homogeneous feature space by using HUDA methods, such as DSFT, GLG, and SFER. Then, we can apply an existing PU learning method to the unlabeled data $\hat{X}_t$ and the positive labeled data $\hat{X}_s$.


However, this naive combination approach can be ineffective due to the following the reason. HUDA methods learn feature transformers that reduce the distribution divergence between $\hat{X}_s$ and $\hat{X}_t$ under some metric, such as MMD. However, recall that $\hat{X}_s$ includes only positive data while $\hat{X}_t$ includes both positive and negative data. Due to this gap, the mapping that reduces the distribution divergence between $\hat{X}_s$ and $\hat{X}_t$ does not necessarily generate a feature space that gives high discriminative performance on the target data, as described in Fig.~\ref{fig:transformation} (top). Indeed, we experimentally confirmed that this naive combination of HUDA and PU learning did not work in Sec.~\ref{experiments}. In the next section, we propose a new method to overcome this difficulty.

\section{Proposed Approach}
To overcome the difficulty described in the previous section, we need to make only the positive target data close to the source data while keeping the negative target data away from the source data in the transformed feature space.

\subsection{Basic Idea}
\label{basic}
Specifically, we regard PU-HUDA as a task to identify likely positive examples from the target data and reduce the distribution divergence between the whole source examples and the likely positive target examples. To solve this task, we exploit adversarial training and thus introduce a discriminator. Hence, our proposed method mainly consists of the following three models as described in Fig.~\ref{fig:proposal} (Details of Fig.~\ref{fig:proposal} are explained in Sec.~\ref{sec:pada}).


\textbf{(1) Feature Transformer} $F:\mathbb{R}^{c+t} \rightarrow \mathbb{R}^s$ transforms the target feature space to that of the source. The target data $X_t$ is transformed into the data $\hat{T}_s=F(X_t)$ in the source-specific feature space via $F$. Thus, we get $\hat{X}_t=[T_c;\hat{T}_s]$ and $X_s$ in the same feature space (see Fig.~\ref{fig:proposal}). Here, we employed an asymmetric transformation \cite{hda} (i.e., only target data is transformed) because it is difficult to train feature transformers for each of the source and target data, which have different label distributions and feature spaces, and therefore it is easy to fall into non-optimal solutions.

\textbf{(2) Classifier} $C:\mathbb{R}^{c+s} \rightarrow \mathbb{R}^2$ identifies positive examples in the transformed unlabeled target data (classifies positive and negative target examples). We represent the output confidence scores of $C$ as $C(\cdot)=(C(\cdot)_0, C(\cdot)_1)$, where $C(\cdot)_1$ and $C(\cdot)_0$ represent the probabilities that the input samples are predicted to be positive and negative, respectively. 

\textbf{(3) Discriminator} $D:\mathbb{R}^{c+s} \rightarrow \mathbb{R}^2$ determines whether the likely positive examples are from the target or source domain (negative examples are predicted as target). The same notation as $C$ applies to $D$ although positive and negative mean source and target for $D$, respectively.

 Since $D$ is trained to discriminate the source and target examples, $F$ and $C$ are trained as follows:   
\begin{itemize}
     \item $F$ is trained so that the source data and the transformed target data predicted by $C$ to be positive cannot be distinguished by $D$. 
    \item $C$ is trained to identify likely positive examples from the target data transformed by $F$ so that they cannot be distinguished from the source data that are all positive by $D$. Therefore, $C$ can classify positive and negative examples.
\end{itemize}
$C$ and $F$ provide feedback to each other. That is, if $C$ can predict the correct positive examples, $F$ can reduce the distribution divergence between the truly positive target data and the source data, and conversely if $F$ can transform the target positive data into the data close to the source data, $C$ can predict the likely positive examples more indistinguishable from the source data. Therefore, by alternately training them with $D$, we can obtain good $C$ and $F$, which are used for binary classification inference on the target data. 

We next consider the optimization ways to train $C$, $D$, and $F$.

\subsection{Naive Optimization Approach}
\label{sec:wada}
We first consider focusing on only positive target data while ignoring negative target data and reducing the distribution divergence between the target positive data and the source data in the transformed feature space.  

The standard objective function used in domain adaptation methods using adversarial learning \cite{dann,adda,dada} to optimize $D$ and $F$ is the following cross-entropy loss:
\begin{align}
\label{strandard}
     \min_{F}\max_{D}V(D, F)
     = &\mathbb{E}_{x_s \sim \mathcal{P}_s(x_s)}[\log D(x_s)_1] \notag \\ 
     &+\mathbb{E}_{x_t \sim \mathcal{P}_t(x_t)}[\log D(\hat{x}_t)_0],
\end{align}
where $\mathcal{P}_s$ and $\mathcal{P}_t$ are data distributions of the source and target data, $\hat{x}_t = [t_c;F(x_t)]$, and $x_t=[t_c;t_t]$. By maximizing Eq.~\ref{strandard}, $D$ is trained to recognize source and target domains, and by minimizing Eq.~\ref{strandard}, $F$ is trained to fool $D$, i.e., to reduce the divergence between the source and target data. 

Since optimizing only Eq.~\ref{strandard} causes the label-gap problem described in Sec.\ref{difficulty}, we incorporate the outputs of $C$ into the second term to focus on the positive target data and train $C$ simultaneously. That is, we optimize the following objective function to train $C$, $F$, and $D$:
\begin{align}
\label{naive}
     \min_{C,F}\max_{D}V_{\rm{wada}}(D, F, &C)
     =  \mathbb{E}_{x_s \sim \mathcal{P}_s(x_s)}[\log D(x_s)_1] \notag \\ 
     &+\frac{\mathbb{E}_{x_t \sim \mathcal{P}_t(x_t)}[C(\hat{x}_t)_1\log D(\hat{x}_t)_0]}{\mathbb{E}_{x_t \sim \mathcal{P}_t(x_t)}[C(\hat{x}_t)_1]}.
\end{align}
By weighting the second term with $C(\cdot)_1$, the training of $D$ and $F$ focuses on the target data that $C$ predicts are likely positive. The denominator of the second term prevents $C$ from outputting 0 for any input. Therefore, by the weighted second term, $C$, $F$, and $D$ are expected to be trained as described in Sec.~\ref{basic}. We call this naive optimization approach \textit{weighted adversarial domain adaptation} (WADA). At first glance, WADA seems to work. However, WADA has the following two problems. 
\begin{itemize}
    \item Since WADA places no restrictions on $C$ incorrectly predicting positive examples to be negative, $C$ will predict as positive only those samples that are sufficiently likely to be positive.
    \item Since the second term in Eq.~\ref{naive} tends to be smaller for negative examples, $F$ will overfit on positive examples, failing to transform negative target data into the features that  $C$ can easily classify.
\end{itemize}

\subsection{Predictive Adversarial Domain Adaptation}
\label{sec:pada}
To address the problems in WADA, we propose \textit{predictive adversarial domain adaptation} (PADA). PADA treats positive and negative examples equally by using an objective function based on KL divergence for confidence scores inspired by PAN \cite{pan}, thus solving the two problems of WADA. The objective function of PADA is as follows:
\begin{align}
\label{pada}
     &\min_{C,F}\max_{D}V_{\rm{pada}}(D, F, C|X_s,X_t) \notag \\
     &= \sum_{i=1}^{n_s}\log D(x_s^i)_1 + \sum_{i=1}^{n_t}\log D(\hat{x}_t^i)_0 \notag \\ 
     &+ \lambda \biggl(\sum_{i=1}^{n_t}KL(D(\hat{x}_t^i)||C(\hat{x}_t^i)) - \sum_{i=1}^{n_t}KL(D(\hat{x}_t^i)||\tilde{C}(\hat{x}_t^i))\biggr),
\end{align}
where $\hat{x}_t^i = [t_c^i;F(x_t^i)]$ and $x_t^i=[t_c^i;t_t^i]$. Each $x_{*}^i$ is a data sample from $X_*$. $\tilde{C}(\cdot)$ denotes the opposite output of $C(\cdot)$, i.e., $\tilde{C}(\cdot)=(C(\cdot)_1, C(\cdot)_0)$.

\textbf{The first and second terms:} cross-entropy for $D$, corresponding to the empirical version of Eq.~\ref{naive}. Optimizing these terms has the following effects on $D$ and $F$.
\begin{itemize}
    \item $D$ can identify the source and target data to some extent. Here, all the target data is regarded as negative for $D$ to give $D$ the ability to recognize the target data.
    \item Since $F$ is trained adversarial to $D$, $F$ can transform the whole target data into the data close to the source data to some extent.
\end{itemize}

\textbf{The third and fourth terms:} KL divergence between the outputs of $C$ and $D$. These terms help to alleviate the label-gap problem caused by using only the first and second terms and the problems in WADA. Since the fourth term only symmetrizes the gradient by the third term and improves performance \cite{pan}, only the effects of the third term are discussed below.
\begin{itemize}
    \item $C$ is trained to output similar scores to those of $D$ for the transformed target data. That is, $C$ tries to give a high score of positivity to the target example that $D$ has difficulty distinguishing from source examples, i.e., the positive target example. $C$ also gives a low score of positivity to the target example that is easy to distinguish from source examples by $D$, i.e., the negative target example. Therefore, $C$ obtains an ability to classify target positive and negative examples.
    
    \item $D$ aims to discriminate source examples and positive target examples. $D$ is trained to output the opposite scores from those of $C$ by the third term. If $C$ identifies the likely positive target example, $D$ tries to predict it as target (negative). Conversely, if $C$ identifies the likely negative target example, $D$ tries to predict it as source (positive). The latter effect is not desirable but is canceled with the effect of the second term.
    
    \item $F$ is trained so that $D$ outputs similar scores to those of $C$, i.e., the likely positive (resp. negative) target examples predicted by $C$ are indistinguishable (resp. distinguishable) from the source examples by $D$. In other words, $F$ tries to make $D$ predict the likely positive (resp. negative) target example predicted by $C$ as source (resp. target). This allows $F$ to transform the positive target data into the data close to the source data while keeping the negative target data away from the source data. The latter is an effect not obtained with WADA. 
\end{itemize}

As a result, we obtain a label-discriminative feature space in the target domain and a good classifier that works on that feature space. Algorithm~\ref{alg:pada} describes the learning algorithm of PADA using stochastic gradient descent.
     
We finally mention the role of common features in PADA. Common features prevent PADA falls into a trivial solution where $C$ predicts all target data to be negative. That is, common features are useful for finding some examples in the target data that are difficult for $D$ to distinguish from the source data, especially in the early stage of learning. It is clear that PADA will not fall into the other trivial solution where $C$ predicts all target data to be positive because $D$ is learned with the target data as negative by the second term of Eq.~\ref{pada}. See the next subsection for how to use common features more explicitly to facilitate the learning of $C$.

\begin{algorithm}[t]
\caption{Learning of PADA by the stochastic gradient descent.}
\label{alg:pada}
\begin{algorithmic}[1]
\REQUIRE positive labeled source data $X_s=[S_c;S_s] \in \mathbb{R}^{n_s\times(c+s)}$, unlabeled target data $X_t=[T_c;T_t] \in \mathbb{R}^{n_t\times(c+t)}$, number of training steps $T$, and initial model parameters $\theta_C, \theta_D, \theta_F$ of $C, D, F$.
\ENSURE \ 
$C, D, F$
\FOR {$T$ steps}
\STATE Sample a mini-batch $\mathbf{x}_s=\{x_s^1, \cdots, x_s^m\}$ from $X_s$ and a mini-batch $\mathbf{x}_t=\{x_t^1, \cdots, x_t^m\}$ from $X_t$.\ \  // training of $D$
\STATE $\theta_D \leftarrow \theta_D + \nabla_{\theta_D} V_{pada}(D, F, C|\mathbf{x}_s,\mathbf{x}_t)$
\STATE Sample a mini-batch $\mathbf{x}_s=\{x_s^1, \cdots, x_s^m\}$ from $X_s$ and a mini-batch $\mathbf{x}_t=\{x_t^1, \cdots, x_t^m\}$ from $X_t$.\ \  // training of $F$
\STATE $\theta_F \leftarrow \theta_F - \nabla_{\theta_F} V_{pada}(D, F, C|\mathbf{x}_s,\mathbf{x}_t)$
\STATE Sample a mini-batch $\mathbf{x}_t=\{x_t^1, \cdots, x_t^m\}$ from $X_t$.\ \  // training of $C$
\STATE $\theta_C \leftarrow \theta_C - \nabla_{\theta_C} V_{pada}(D, F, C|\ \cdot\ ,\mathbf{x}_t)$
\ENDFOR
\STATE Return $C, D, F$.
\end{algorithmic}
\end{algorithm}

\subsection{Soft labeling mechanism to improve PADA}
In the early stage of learning PADA, since $F$ is not fully trained, the features $\hat{T}_s$ created by $F$ may make the whole features $\hat{X}_t=[T_c;\hat{T}_s]$ useless. As a result, $C$ might fail to identify likely positive examples from $\hat{X}_t$, which leads to the unstable learning of $C$ and $F$. Therefore, in this section, we consider guiding the learning of $C$ by a \textit{base classifier} that can be trained stably using only common features without $F$.

First, we apply an existing PU learning method to the common features of the source data and target data, $S_c$ and $T_c$. Let $C_0$ be the base classifier obtained from the existing PU learning method using $S_c$ and $T_c$ as the positive and unlabeled data, respectively. If the common features contain some useful information for classification, $C_0$ should attain some degree of classification accuracy. Therefore, we can use $C_0$ to guide the learning of $C$. By using $C_0$, the objective of PADA is modified as follows:

\begin{align}
\label{pada2}
     &\min_{C,F}\max_{D}V_{\rm{pada}}'(D, F, C|X_s,X_t) \notag \\
     &= V_{\rm{pada}}(D, F, C|X_s,X_t) \notag \\
     &+ \eta \biggl(\sum_{i=1}^{n_t}KL(C_0(t_c^i)||C(\hat{x}_t^i)) - \sum_{i=1}^{n_t}KL(C_0(t_c^i)||\tilde{C}(\hat{x}_t^i))\biggr),
\end{align}
where $\eta$ is a balancing parameter. Note that $C_0$ takes only common features as inputs. The second and third terms are added to the original objective of PADA. Since the third term is a symmetrization term,  we discuss only the second term. In the second term, $C_0$ gives soft labels\footnote{Soft labels are usually used in the context of \textit{distillation} \cite{distillation}, which transfers the knowledge of large models to small models.} to the unlabeled data $\hat{X}_t$, and $C$ is trained using them as teacher labels. By this term, the learning of $C$ will be accelerated when $D$ and $F$ are not yet learned enough, which also leads to accelerated learning of $D$ and $F$ as a result. 

Once the learning of $C$ is finished, $C$ is expected to perform better than $C_0$. Therefore, it is also expected that training PADA again using this $C$ as a base classifier $C_0$ will yield a better classifier. With this insight, we adopt repeating this soft-labeling process until the accuracy of the finally produced classifier saturates. We note that, in the second and later rounds, $C_0$ can take both common and target-specific features as inputs. 

\section{Experiments}
\label{experiments}
\subsection{Experimental Setup}
\subsubsection{Datasets}
We reorganized three datasets and used them to evaluate our proposed methods since no public dataset was directly related to the PU-HUDA setting. Detailed information about the datasets is presented in Table~\ref{dataset}. 

\begin{table*}
\centering
\caption{Detailed information about the processed datasets, Movielens-Netflix, 20-Newsgroups, and Default of credit card clients.}\label{dataset}
\begin{tabular}{cc|c|c|c|c|c|c|c|c}
\hline
\multicolumn{2}{c|}{\multirow{4}{*}{\textbf{Dataset}}} &  \multicolumn{5}{|c|}{\textbf{Number of samples}} & 
\multicolumn{3}{|c}{\textbf{Number of features}}\\
\cline{3-10}
& & \multicolumn{3}{|c|}{Training} & \multicolumn{2}{|c|}{Testing} & & \multicolumn{2}{|c}{Domain-specific} \\
\cline{3-7}
\cline{9-10}
& & \multicolumn{2}{|c|}{Target} & Source & \, \multirow{2}{*}{Pos} \, & \, \multirow{2}{*}{Neg} \, & Common & \ \multirow{2}{*}{Target} \ & \multirow{2}{*}{Source} \\
\cline{3-5}
& & \; Pos \; & \; Neg \; & Pos & & & & & \\
\hline
\multicolumn{2}{c|}{Movielens-Netflix} & 6,836 & 9,164 & 8,835 & 55,725  & 74,816 & 4 & 13 & 20 \\ 
\cline{1-10}
\multirow{6}{*}{20-Newsgroups} & CR & 1,168 & 1,197 & 1,768 & 397  & 390 & \multirow{6}{*}{500} & \multirow{6}{*}{5,000} & \multirow{6}{*}{5,000} \\
& CS & 1,168 & 1,187 & 1,768 & 378  & 406 & & & \\
& CT & 1,168 & 842 & 1,768 & 391  & 278 & & & \\
& RS & 1,192 & 1,187 & 1,197 & 396  & 396 & & & \\
& RT & 1,192 & 842 & 1,197 & 412  & 266 & & & \\
& ST & 1,187 & 842 & 1,186 & 380  & 296 & & & \\
\cline{1-10}
\multirow{3}{*}{Credit card} & 0.5 & 3,763 & 3,763 & 2,873 & 753  & 753 & \multirow{3}{*}{4} & \multirow{3}{*}{13} & \multirow{3}{*}{13} \\
& 0.3 & 3,763 & 8,780 & 2,873 & 732 & 1,778 & & & \\
& 0.1 & 1,594 & 14,349 & 2,873 & 325  & 2,865 & & & \\
\hline
\end{tabular}
\end{table*}

\textit{Movielens-Netflix} \cite{movielens,netflix}: Movielens and Netflix are two movie datasets that contain users and their ratings for movies on a scale of [0,5]. To create a classification dataset using them, we focus on the movie genre rather than the movie title. We first calculate each user's average rating for each genre, and the differences between those and the user's average rating for all movies are used as features. The genres common to Movielens and Netflix correspond to common features. We select one feature from common features and label each user as positive or negative according to whether the value for the feature is positive or negative, i.e., whether the user likes the corresponding genre or not. We chose Adventure as the label because the correlation of Adventure with other features is high, and the ratios of positive and negative are approximately equal. Since the default proportion of common genres is large, we randomly chose 4 genres from the 16 common genres and used them as common features. Half of the remaining common genres were used for Movielens-specific features and half for Netflix-specific features. We assigned Movielens as the target and Netflix as the source.

\textit{20-Newsgroups} \cite{20-newsgroups}: a collection of about 20,000 documents belonging to four top categories: computer (C), recording (R), science (S), and talk (T), each containing four more subcategories. We create a binary classification task as in DSFT \cite{dsft} by selecting two top categories and assigning them positive and negative labels. Half of the subcategories are used for source and half for target. Each document is represented by tf-idf features. We use the top 500 most frequent common words in the source and target domain as common features and the top 5000 frequent specific words in each domain as domain-specific features.

\textit{Default of credit card clients} \cite{credit}: a collection of credit card records, including user demographics, history of payments, bill statements, and previous payments, with users’ default payments as binary labels. We assign male users as source and female users as target. User demographics are used as common features, which may often occur in reality. We use bill statements and half of the history of payments as source-specific features, and previous payments and half of the history of payments as target-specific features. We vary the positive ratio of the target data from 0.5 to 0.1 to see the sensitivity of PADA to it.

\subsubsection{Compared methods}
We compare the performances of \textbf{PADA} and PADA with soft-labeling ($\rm{\textbf{PADA}}_\textit{\textbf{S}}$) to the three types of baseline methods in addition to \textbf{WADA}. 

The first baseline is to apply existing PU learning methods to only common features. As existing PU learning methods, we use \textbf{NaivePU}, which treats the entire target data as negative, \textbf{nnPU} \cite{nnpu}, and \textbf{PAN} \cite{pan}.

The second type of baseline is \textbf{DIST}, which uses not only common features but also target-specific features\footnote{Note that DIST can not transfer the knowledge of the source-specific features to the target domain.}. DIST only uses the soft labeling mechanism of $\rm{PADA}_\textit{S}$, i.e., first gives soft labels to the unlabeled target data using the model trained by the first baseline and trains a classifier using all the target features and the soft labels as in usual distillation \cite{distillation}. As the first baseline, we adopt PAN because of its high performance. 

The last baseline is a direct combination of existing HUDA and PU learning methods described in Sec.~\ref{difficulty}. This baseline first constructs a homogeneous feature space by existing HUDA methods and next applies PU learning methods to the source and target data on the constructed homogeneous feature space. As a PU learning method, we adopt PAN for the same reason as DIST. We use two HUDA methods as baselines, \textbf{DSFT} \cite{dsft} and \textbf{SFER} \cite{sfer}. The linear and non-linear DSFT are denoted by $\rm{\textbf{DSFT}}_\textit{\textbf{l}}$ and $\rm{\textbf{DSFT}}_\textit{\textbf{nl}}$, respectively. SFER is not used for 20-Newsgroups because SFER can only be applied to low dimensional features due to computational cost.  


\subsubsection{Implementation details} The binary classifier trained by each method was tested on the target domain. We split each target data into training, validation, and testing datasets. Each method was tuned by the validation dataset and tested by the testing dataset. We implemented each method three times and used average test accuracy as the test metric. For PADA and WADA, we selected the best learning rate in [0.0001, 0.0005, 0.001, 0.005, 0.01, 0.05, 0.1], $\lambda$ and $\eta$ in [0.0001, 0.0005, 0.001, 0.005, 0.01, 0.05, 0.1, 0.5, 1]. We repeated the soft-labeling of PADA until the convergence of validation accuracy. The batch size of PADA was set to 128. For a fair comparison, each method adopted a linear classifier and a linear discriminator, i.e., logistic regressions. We also used a linear transformation for a feature transformer. 

\subsection{Results and Analysis}
\subsubsection{Performance comparison}

\begin{table*}[t]
\centering
\caption{Performance comparison using accuracy (\%) for Movivlens-Netflix dataset. Four common features were randomly selected five times, which makes 5 experimental settings.}\label{main1}
\begin{tabular}{c|c|c|c|c|c|c|c|c|c|c}
\hline
\multirow{2}{*}{\textbf{Settings}} &  \multicolumn{10}{c}{\textbf{Methods}}\\
\cline{2-11}
  & \, NaivePU \, & \, nnPU \, & \, PAN \, & \; DIST \; &$\, \rm{DSFT}_{\textit{l}}$\, & $\rm{DSFT}_{\textit{nl}}$ & SFER &\, WADA \, &\,  PADA \, & $\rm{PADA}_\textit{S}$ \\
\hline
A & 59.01 & 57.89 & 59.93 & 62.64 & 62.34 & 57.33 & 57.84 & 64.40 & \textbf{66.29} & 66.13 \\
B & 58.40 & 57.55 & 59.99 & 62.50 & 57.09 & 58.55 & 60.53 & 61.49 & 61.90 & \textbf{63.14} \\
C & 63.72 & 63.10 & 62.64 & 65.07 & 68.35 & 60.94 & 57.82 & 67.90 & 67.82 & \textbf{68.93} \\
D & 62.93 & 63.26 & 65.49 & 66.85 & 69.17 & 58.71 & 62.99 & 65.52 & \textbf{70.26} & 69.70 \\
E & 62.65 & 63.10 & 66.61 & 66.59 & 66.34 & 61.54 & 57.70 & 64.60 & 65.44 & \textbf{66.72} \\
\hline
\hline
Average & 61.34 & 60.98 & 62.93 & 64.73 & 64.66 & 59.42 & 59.38 & 64.78 & 66.34 & \textbf{66.92} \\
\hline
\end{tabular}
\end{table*}

\begin{table*}[t]
\centering
\caption{Performance comparison using accuracy (\%) for 20-Newsgroups dataset. We selected two top categories and assigned them as positive and negative labels.}\label{main2}
\begin{tabular}{cc|c|c|c|c|c|c|c|c|c|c}
\hline
\multicolumn{2}{c|}{\textbf{Labels}} &  \multicolumn{10}{c}{\textbf{Methods}}\\
\cline{3-12}
 \textbf{Pos.} & \textbf{Neg.} & \, NaivePU \, & \, nnPU \, & \, PAN \, & \; DIST \; &$\, \rm{DSFT}_{\textit{l}}$\, & $\rm{DSFT}_{\textit{nl}}$ & SFER & \, WADA\, &\,  PADA \, & $\rm{PADA}_\textit{S}$ \\
\hline
C & R & 63.79 & 73.10 & 80.14 & 84.88 & 84.16 & 83.02 & - & 78.37 & 80.43 & \textbf{86.11} \\
C & S & 57.82 & 59.91 & 64.71 & 64.75 & 66.07 & \textbf{69.81} & - & 57.63 & 65.48 & 66.54 \\
C & T & 57.85 & 73.84 & 80.67 & 79.42 & 80.17 & 80.47 & - & 77.18 & 80.67 & \textbf{83.76} \\
R & S & 51.43 & 50.80 & 50.93 & 50.13 & 55.72 & 51.35 & - & 53.75 & 55.81 & \textbf{56.73} \\
R & T & 57.18 & 58.21 & 66.37 & 66.52 & 65.44 & 67.01 & - & 60.28 & 66.96 & \textbf{67.21} \\
S & T & 52.76 & 54.19 & 66.47 & 67.06 & 65.78 & 66.17 & - & 56.14 & 66.91 & \textbf{71.40} \\
\hline
\hline
\multicolumn{2}{c|}{Average} & 56.80 & 61.68 & 68.21 & 68.79 & 69.57 & 69.64 & - & 63.89 & 69.38 & \textbf{71.96} \\
\hline
\end{tabular}
\end{table*}

\begin{table*}[t]
\centering
\caption{Performance comparison using accuracy (\%) and AUC (\%) for Default of credit card clients dataset. The positive ratio of target data was varied from 0.5 to 0.1.}\label{main3}
\begin{tabular}{cc|c|c|c|c|c|c|c|c|c|c}
\hline
\textbf{Positive} & \textbf{Test} & \multicolumn{10}{c}{\textbf{Methods}}\\
\cline{3-12}
 \textbf{ratio} & \textbf{metric} & \, NaivePU \, & \, nnPU \, & \, PAN \, & \; DIST \; &$\, \rm{DSFT}_{\textit{l}}$\, & $\rm{DSFT}_{\textit{nl}}$ & SFER &\, WADA \, &\,  PADA \, & $\rm{PADA}_\textit{S}$ \\
\hline
\multirow{2}{*}{0.5} & Accuracy & 54.58 & 55.47 & 54.60 & 57.85 & 44.42 & 54.07 & 49.40 & 59.88 & 60.09 & 63.70 \\
 & AUC & 58.03 & 58.21 & 58.76 & 61.72 & 37.75 & 55.93 & 48.94 & 65.25 & 63.02 & 68.42 \\
\multirow{2}{*}{0.3} & Accuracy & 61.29 & 69.63 & 62.79 & 66.02 & 58.91 & 65.42 & 70.57 & 69.63 & 65.31 & 68.53 \\
 & AUC & 58.76 & 60.09 & 61.68 & 66.04 & 49.33 & 45.88 & 58.73 & 67.79 & 65.55 & 68.09 \\
\multirow{2}{*}{0.1} & Accuracy & 64.03 & 83.78 & 63.32 & 67.64 & 62.87 & 82.59 & 80.61 & 69.20 & 70.73 & 72.20 \\
 & AUC & 59.46 & 58.24 & 61.24 & 66.74 & 70.52 & 68.51 & 36.85 & 63.03 & 65.77 & 61.93 \\
\hline
\end{tabular}
\end{table*}

The test accuracy results of all compared methods are reported in Table~\ref{main1}, Table~\ref{main2}, and Table~\ref{main3}. We also report AUC for Default of credit card clients because the positive ratio is changed to small. The best result among compared methods is written in bold. 

\textit{Movielens-Netflix:} Table~\ref{main1} shows the results for Movielens-Netflix and we can see that PADA and $\rm{PADA}_\textit{S}$ outperform all the baseline methods in every setting. Specifically, they improve the first baselines such as PAN using only common features by about 4\% on average and by about 6\% in a particular setting, which is an effect of source-specific and target-specific features. Note that DIST also improves PAN. This is because the generalization performance is improved by using not only common features but also target-specific features when performing distillation by the soft labels given by PAN. Moreover, PADA outperforms DIST, which indicates that they successfully transfers the source-specific features to the target domain. Compared to PADA, WADA's performance is modest, as expected. We can also observe the positive effect of the soft-labeling by $\rm{PADA}_\textit{S}$ over PADA. We also note that the naive HUDA and PU learning combination methods, such as $\rm{DSFT}_\textit{l}$, $\rm{DSFT}_\textit{nl}$, and SFER, provide limited performances. Unlike DSFT, which does nothing for common features, SFER, which transforms all features, does not work at all. This fact is consistent with the difficulty of these methods described in Sec.~\ref{problem}. See Sec.~\ref{sec:adaptation} for verification of the effectiveness of PADA by observing the distribution divergence between the positive (or negative) target and source data in the transformed feature space.

\textit{20-Newsgroups:} The results for 20-Newsgroups are presented in Table~\ref{main2}. In this dataset, we can observe the positive effect of DSFT. In the setting C-S, for example, $\rm{DSFT}_\textit{nl}$ improves PAN by 5\%. Overall, however, its improvement over PAN is modest. Although the improvement of PADA over PAN is also modest, the soft-labeling of $\rm{PADA}_\textit{S}$ is considerably effective in this dataset, outperforming the baseline methods in almost all settings. We note that at most three iterations of soft labeling in $\rm{PADA}_\textit{S}$ were sufficient. We finally mention that WADA does not work well at all in this dataset. 

\textit{Default of credit card clients:} Table~\ref{main3} shows the results for Default of credit card clients. When the positive ratio is 0.5, we can see that $\rm{PADA}_\textit{S}$ is the best method in terms of both accuracy and AUC. When the positive ratio is 0.3, WADA outperforms PADA although $\rm{PADA}_\textit{S}$ also shows almost equivalent performance to WADA. Note that the accuracy of SFER is high, its AUC is low, which means it predicts most of the data as negative. In the case of 0.1 positive ratio, PADA and $\rm{PADA}_\textit{S}$ also show the improvements over most of the baselines. However, $\rm{DSFT}_\textit{nl}$ is better than them. As a result, DSFT may be effective in particular datasets and settings, but its effective range is very limited.

\subsubsection{Analysis on the number of common features}

\begin{figure}[t]
    \centering
    \includegraphics[width=0.45\textwidth]{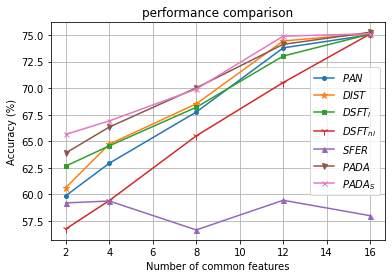}
    \caption{Effect of the number of selected common features for Movielens-Netflix} \label{fig:experiment}
    \vspace{-1\baselineskip}
\end{figure}

Using the Movielens-Netflix dataset, we further investigate the effect of the number of common features on our proposed methods and several baseline methods, including PAN, DIST, DSFT, and SFER. Fig.~\ref{fig:experiment} shows the test accuracy when varying the number of selected common features from 2 to 16 (all genres common to Movielens and Netflix). We can see that the performance of each method degrades as the number of common features decreases, but $\rm{PADA}_\textit{S}$ shows the least degradation. Conversely, the greater the number of common features, the smaller the performance difference between all methods except SFER. This is because as the number of common features increases, the number of target-specific features decreases, i.e., the target-specific features lose their usefulness. SFER, a naive combination of HUDA and PU learning is not valid regardless of the number of common features, which means that the homogeneous feature spaces obtained by it lose or inhibit the usefulness of common features.



\subsection{Adaptation analysis of PADA}
\label{sec:adaptation}
In this section, we experimentally confirm that PADA transforms the positive target data into data close to the source data while keeping the negative target data away from the source data using the Movielens-Netflix dataset. 

We compare PADA with three baselines: (1) Common, which uses only the raw common features of the source and target data, (2) DSFT, a naive combination of HUDA and PU learning methods, and (3) WADA. To see the distribution distance between the source data and the positive (resp. negative) target data in the resulting feature space, we define a metric $Acc_d(s,t_p)$ (resp. $Acc_d(s,t_n)$), the test accuracy of discrimination by a \textit{test discriminator} that discriminates the source data and the positive (resp. negative) target data. The test discriminator is trained on the resulting feature space after finishing the learning of each method or only on the common feature space for Common. Note that the test discriminator is different from the discriminator trained for PADA and WADA. The true labels of the target data are used to train the test discriminator. If $Acc_d(s,t_p)$ (resp. $Acc_d(s,t_n)$) is close to 0.5, the distribution divergence between the source data and the positive (resp. negative) target data is considered small. Therefore, small $Acc_d(s,t_p)$ and large $Acc_d(s,t_n)$ are desirable in PU-HUDA.

The results are presented in Table~\ref{adaptation}. As expected, we can see that $Acc_d(s,t_p)$ of PADA is smaller than those of the other methods, while $Acc_d(s,t_n)$ of PADA is larger than it. This means that the positive target data are transformed into data close to the source data while keeping the distance between the negative target data and the source data. $Acc_d(s,t_p)$ of Common is smaller than that of PADA, but $Acc_d(s,t_n)$ of Common is also small. As a result, the accuracy of the existing PU learning methods using only common features is low. Even for DSFT, the difference between $Acc_d(s,t_p)$ and $Acc_d(s,t_n)$ is small, which is consistent with the problem of the naive combination approaches described in Sec.~\ref{difficulty}. We also note that although the difference between $Acc_d(s,t_p)$ and $Acc_d(s,t_n)$ of WADA is large, its $Acc_d(s,t_p)$ is also large. Thus, if WADA is trained to make $Acc_d(s,t_p)$ smaller, $Acc_d(s,t_n)$ is expected to be even closer to $Acc_d(s,t_p)$, which is an undesirable result as described in Sec.~\ref{sec:wada}.

\section{Conclusion}
We defined a new heterogeneous domain adaptation task where only positive examples are available in the source domain. To overcome the gap in label distribution between the source and target data, we proposed a novel method that integrates the feature alignment and PU learning in a unified adversarial training framework. Moreover, we experimentally showed that our proposed method outperformed several baseline methods. 

Finally, it should be mentioned that this paper does not take into account the divergence between the distributions of the source common features and target common features. It is a future work to reduce the divergence between the source common features and target common features (if there is any divergence) and increase the performance of our proposed method.



\begin{table}[t]
\centering
\caption{The average test accuracy (\%) of the test discriminator for each method using Mevielens-Netflix.}\label{adaptation}
\begin{tabular}{c|c|c|c}
\hline
\multirow{2}{*}{\textbf{Method}} & \multirow{2}{*}{\textbf{Accuracy of method}} & \multicolumn{2}{|c}{\textbf{Accuracy of discrimination}}\\
\cline{3-4}
& & $Acc_{d}(s,t_p)$ & $Acc_{d}(s,t_n)$ \\
\hline
Common & 62.93 & 54.79 & 55.79 \\
DSFT & 64.73 & 72.59 & 76.58 \\
WADA & 64.78 & 70.73 & 77.12 \\
PADA & 66.34 & 61.08 & 66.02 \\
\hline
\end{tabular}
\end{table}


\bibliographystyle{IEEEtran}
\bibliography{IEEEexample}

\begin{thebibliography}{10}
\providecommand{\url}[1]{#1}
\csname url@samestyle\endcsname
\providecommand{\newblock}{\relax}
\providecommand{\bibinfo}[2]{#2}
\providecommand{\BIBentrySTDinterwordspacing}{\spaceskip=0pt\relax}
\providecommand{\BIBentryALTinterwordstretchfactor}{4}
\providecommand{\BIBentryALTinterwordspacing}{\spaceskip=\fontdimen2\font plus
\BIBentryALTinterwordstretchfactor\fontdimen3\font minus
  \fontdimen4\font\relax}
\providecommand{\BIBforeignlanguage}[2]{{%
\expandafter\ifx\csname l@#1\endcsname\relax
\typeout{** WARNING: IEEEtran.bst: No hyphenation pattern has been}%
\typeout{** loaded for the language `#1'. Using the pattern for}%
\typeout{** the default language instead.}%
\else
\language=\csname l@#1\endcsname
\fi
#2}}
\providecommand{\BIBdecl}{\relax}
\BIBdecl

\bibitem{bias}
A.~Torralba and A.~A. Efros, ``Unbiased look at dataset bias,'' in \emph{CVPR
  2011}, 2011, pp. 1521--1528.

\bibitem{transfer}
S.~J. Pan and Q.~Yang, ``A survey on transfer learning,'' \emph{IEEE
  Transactions on Knowledge and Data Engineering}, vol.~22, no.~10, pp.
  1345--1359, 2010.

\bibitem{deepda}
M.~Wang and W.~Deng, ``Deep visual domain adaptation: A survey,''
  \emph{Neurocomputing}, vol. 312, pp. 135--153, 2018.

\bibitem{hda}
O.~Day and T.~M. Khoshgoftaar, ``A survey on heterogeneous transfer learning,''
  \emph{Journal of Big Data}, vol.~4, 2017.

\bibitem{kcca}
Y.-R. Yeh, C.-H. Huang, and Y.-C.~F. Wang, ``Heterogeneous domain adaptation
  and classification by exploiting the correlation subspace,'' \emph{IEEE
  Transactions on Image Processing}, vol.~23, no.~5, pp. 2009--2018, 2014.

\bibitem{hhtl}
J.~Zhou, S.~Pan, I.~Tsang, and Y.~Yan, ``Hybrid heterogeneous transfer learning
  through deep learning,'' \emph{Proceedings of the AAAI Conference on
  Artificial Intelligence}, vol.~28, no.~1, 2014.

\bibitem{dsft}
P.~Wei, Y.~Ke, and C.~K. Goh, ``A general domain specific feature transfer
  framework for hybrid domain adaptation,'' \emph{IEEE Transactions on
  Knowledge and Data Engineering}, vol.~31, no.~8, pp. 1440--1451, 2019.

\bibitem{ot}
T.~Aritake and H.~Hino, ``Domain adaptation with optimal transport for extended
  variable space,'' in \emph{2022 International Joint Conference on Neural
  Networks (IJCNN)}, 2022, pp. 1--9.

\bibitem{sfer}
F.~Liu, J.~Lu, and G.~Zhang, ``Unsupervised heterogeneous domain adaptation via
  shared fuzzy equivalence relations,'' \emph{IEEE Transactions on Fuzzy
  Systems}, vol.~26, no.~6, pp. 3555--3568, 2018.

\bibitem{glg}
F.~Liu, G.~Zhang, and J.~Lu, ``Heterogeneous domain adaptation: An unsupervised
  approach,'' \emph{IEEE Transactions on Neural Networks and Learning Systems},
  vol.~31, no.~12, pp. 5588--5602, 2020.

\bibitem{pu}
J.~Bekker and J.~Davis, ``Learning from positive and unlabeled data: A
  survey,'' \emph{Machine Learning}, vol. 109, pp. 719--760, 2020.

\bibitem{gan}
I.~Goodfellow, J.~Pouget-Abadie, M.~Mirza, B.~Xu, D.~Warde-Farley, S.~Ozair,
  A.~Courville, and Y.~Bengio, ``Generative adversarial nets,'' in
  \emph{Advances in Neural Information Processing Systems}, Z.~Ghahramani,
  M.~Welling, C.~Cortes, N.~Lawrence, and K.~Weinberger, Eds., vol.~27.\hskip
  1em plus 0.5em minus 0.4em\relax Curran Associates, Inc., 2014.

\bibitem{pan}
W.~Hu, R.~Le, B.~Liu, F.~Ji, J.~Ma, D.~Zhao, and R.~Yan, ``Predictive
  adversarial learning from positive and unlabeled data,'' \emph{Proceedings of
  the AAAI Conference on Artificial Intelligence}, vol.~35, no.~9, pp.
  7806--7814, 2021.

\bibitem{pivot}
P.~Prettenhofer and B.~Stein, ``Cross-language text classification using
  structural correspondence learning,'' in \emph{Proceedings of the 48th Annual
  Meeting of the Association for Computational Linguistics}.\hskip 1em plus
  0.5em minus 0.4em\relax Uppsala, Sweden: Association for Computational
  Linguistics, 2010, pp. 1118--1127.

\bibitem{fsr}
K.~D. Feuz and D.~J. Cook, ``Transfer learning across feature-rich
  heterogeneous feature spaces via feature-space remapping (fsr),'' \emph{ACM
  Trans. Intell. Syst. Technol.}, vol.~6, no.~1, 2015.

\bibitem{two-steps1}
B.~Liu, W.~S. Lee, P.~S. Yu, and X.~Li, ``Partially supervised classification
  of text documents,'' in \emph{Proceedings of the Nineteenth International
  Conference on Machine Learning}, ser. ICML '02.\hskip 1em plus 0.5em minus
  0.4em\relax San Francisco, CA, USA: Morgan Kaufmann Publishers Inc., 2002, p.
  387–394.

\bibitem{two-steps2}
H.~Yu, J.~Han, and K.~C.-C. Chang, ``Pebl: Positive example based learning for
  web page classification using svm,'' in \emph{Proceedings of the Eighth ACM
  SIGKDD International Conference on Knowledge Discovery and Data Mining}, ser.
  KDD '02.\hskip 1em plus 0.5em minus 0.4em\relax New York, NY, USA:
  Association for Computing Machinery, 2002, p. 239–248.

\bibitem{noise1}
B.~Liu, Y.~Dai, X.~Li, W.~Lee, and P.~Yu, ``Building text classifiers using
  positive and unlabeled examples,'' in \emph{Third IEEE International
  Conference on Data Mining}, 2003, pp. 179--186.

\bibitem{noise2}
C.~Gong, H.~Shi, T.~Liu, C.~Zhang, J.~Yang, and D.~Tao, ``Loss decomposition
  and centroid estimation for positive and unlabeled learning,'' \emph{IEEE
  Transactions on Pattern Analysis and Machine Intelligence}, vol.~43, no.~3,
  pp. 918--932, 2021.

\bibitem{upu}
M.~D. Plessis, G.~Niu, and M.~Sugiyama, ``Convex formulation for learning from
  positive and unlabeled data,'' in \emph{Proceedings of the 32nd International
  Conference on Machine Learning}, ser. Proceedings of Machine Learning
  Research, F.~Bach and D.~Blei, Eds., vol.~37.\hskip 1em plus 0.5em minus
  0.4em\relax Lille, France: PMLR, 2015, pp. 1386--1394.

\bibitem{nnpu}
R.~Kiryo, G.~Niu, M.~C. du~Plessis, and M.~Sugiyama, ``Positive-unlabeled
  learning with non-negative risk estimator,'' in \emph{Advances in Neural
  Information Processing Systems}, I.~Guyon, U.~V. Luxburg, S.~Bengio,
  H.~Wallach, R.~Fergus, S.~Vishwanathan, and R.~Garnett, Eds., vol.~30.\hskip
  1em plus 0.5em minus 0.4em\relax Curran Associates, Inc., 2017.

\bibitem{puda}
J.~Sonntag, G.~Behrens, and L.~Schmidt-Thieme, ``Positive-unlabeled domain
  adaptation,'' in \emph{2022 IEEE 9th International Conference on Data Science
  and Advanced Analytics (DSAA)}, 2022, pp. 1--10.

\bibitem{plt1}
P.~Mignone and G.~Pio, ``Positive unlabeled link prediction via transfer
  learning for gene network reconstruction,'' in \emph{Foundations of
  Intelligent Systems}, M.~Ceci, N.~Japkowicz, J.~Liu, G.~A. Papadopoulos, and
  Z.~W. Ra{\'{s}}, Eds.\hskip 1em plus 0.5em minus 0.4em\relax Cham: Springer
  International Publishing, 2018, pp. 13--23.

\bibitem{plt2}
P.~Mignone, G.~Pio, and M.~Ceci, ``Distributed heterogeneous transfer learning
  for link prediction in the positive unlabeled setting,'' in \emph{2022 IEEE
  International Conference on Big Data (Big Data)}, 2022, pp. 5536--5541.

\bibitem{osda1}
M.~R. Loghmani, M.~Vincze, and T.~Tommasi, ``Positive-unlabeled learning for
  open set domain adaptation,'' \emph{Pattern Recognition Letters}, vol. 136,
  pp. 198--204, 2020.

\bibitem{osda2}
S.~Garg, S.~Balakrishnan, and Z.~C. Lipton, ``Domain adaptation under open set
  label shift,'' in \emph{Advances in Neural Information Processing Systems},
  A.~H. Oh, A.~Agarwal, D.~Belgrave, and K.~Cho, Eds., 2022.

\bibitem{dann}
Y.~Ganin, E.~Ustinova, H.~Ajakan, P.~Germain, H.~Larochelle, F.~Laviolette,
  M.~Marchand, and V.~Lempitsky, ``Domain-adversarial training of neural
  networks,'' \emph{The journal of machine learning research}, vol.~17, no.~1,
  pp. 2096--2030, 2016.

\bibitem{adda}
E.~Tzeng, J.~Hoffman, K.~Saenko, and T.~Darrell, ``Adversarial discriminative
  domain adaptation,'' in \emph{Proceedings of the IEEE Conference on Computer
  Vision and Pattern Recognition (CVPR)}, 2017.

\bibitem{dada}
H.~Tang and K.~Jia, ``Discriminative adversarial domain adaptation,''
  \emph{Proceedings of the AAAI Conference on Artificial Intelligence},
  vol.~34, no.~04, pp. 5940--5947, 2020.

\bibitem{distillation}
G.~E. Hinton, O.~Vinyals, and J.~Dean, ``Distilling the knowledge in a neural
  network,'' \emph{ArXiv}, vol. abs/1503.02531, 2015.

\bibitem{movielens}
\BIBentryALTinterwordspacing
``Movielens,'' 2019. [Online]. Available:
  \url{https://grouplens.org/datasets/movielens/}
\BIBentrySTDinterwordspacing

\bibitem{netflix}
\BIBentryALTinterwordspacing
``Netflix prize data,'' 2009. [Online]. Available:
  \url{https://www.kaggle.com/datasets/netflix-inc/netflix-prize-data}
\BIBentrySTDinterwordspacing

\bibitem{20-newsgroups}
K.~Lang, ``Newsweeder: Learning to filter netnews,'' in \emph{Machine Learning
  Proceedings 1995}, A.~Prieditis and S.~Russell, Eds.\hskip 1em plus 0.5em
  minus 0.4em\relax San Francisco (CA): Morgan Kaufmann, 1995, pp. 331--339.

\bibitem{credit}
I.-C. Yeh and C.~hui Lien, ``The comparisons of data mining techniques for the
  predictive accuracy of probability of default of credit card clients,''
  \emph{Expert Systems with Applications}, vol.~36, no. 2, Part 1, pp.
  2473--2480, 2009.

\end{thebibliography}

\end{document}